\documentclass[conference]{IEEEtran}
\IEEEoverridecommandlockouts
\usepackage{comment}
\usepackage{cite}
\usepackage{amsmath,amssymb,amsfonts}
\usepackage{algorithmic}
\usepackage{graphicx}
\usepackage{textcomp}
\usepackage{xcolor}
\usepackage{array}
\def\BibTeX{{\rm B\kern-.05em{\sc i\kern-.025em b}\kern-.08em
    T\kern-.1667em\lower.7ex\hbox{E}\kern-.125emX}}

\makeatletter
\newcommand{\linebreakand}{%
  \end{@IEEEauthorhalign}
  \hfill\mbox{}\par
  \mbox{}\hfill\begin{@IEEEauthorhalign}
}
\makeatother
    
\begin{document}

\title{Preliminary Prototyping of Avoidance Behaviors Triggered by a User's Physical Approach to a Robot}

\author{\IEEEauthorblockN{Tomoko Yonezawa}
\IEEEauthorblockA{\textit{Faculty of Informatics.}\\
\textit{Kansai University.}\\
Osaka, Japan\\
yone@kansai-u.ac.jp}
\and
\IEEEauthorblockN{
Hirotake Yamazoe}
\IEEEauthorblockA{\textit{Graduate School of Engineering.} \\
\textit{University of Hyogo.}\\
Hyogo, Japan \\
yamazoe@eng.u-hyogo.ac.jp}
\and
\IEEEauthorblockN{
Atsuo Fujino}
\IEEEauthorblockA{\textit{Graduate School of Informatics.} \\
\textit{Kansai University.}\\
Osaka, Japan.}
\linebreakand
\IEEEauthorblockN{
Daigo Suhara}
\IEEEauthorblockA{\textit{Graduate School of Informatics.} \\
\textit{Kansai University.}\\
Osaka, Japan.}
\and
\IEEEauthorblockN{
Takaya Tamamoto}
\IEEEauthorblockA{\textit{Graduate School of Informatics.} \\
\textit{Kansai University.}\\
Osaka, Japan.}
\and
\IEEEauthorblockN{
Yuto Nishiguchi}
\IEEEauthorblockA{\textit{Graduate School of Informatics.} \\
\textit{Kansai University.}\\
Osaka, Japan}
}

\maketitle

\begin{abstract}
Human-robot interaction frequently involves physical proximity or contact. In human-human settings, people flexibly accept, reject, or tolerate such approaches depending on the relationship and context. We explore the design of a robot's rejective internal state and corresponding avoidance behaviors, such as withdrawing or pushing away, when a person approaches. We model the accumulation and decay of discomfort as a function of interpersonal distance, and implement tolerance (endurance) and limit-exceeding avoidance driven by the Dominance axis of the PAD affect model. The behaviors and their intensities are realized on an arm robot. Results illustrate a coherent pipeline from internal state parameters to graded endurance motions and, once a limit is crossed, to avoidance actions.
\end{abstract}


\section{Introduction}
Human-robot interaction (HRI) should be designed with social context in mind, especially in scenarios that involve physical proximity or contact. While much robot design has emphasized positive affect---such as affinity, trust, and likability---intimate, person-to-person encounters should not assume unconditional acceptance by the robot. Designers should also consider negative stances such as rejection and avoidance. Such stances are valuable both for safeguarding the robot (in the spirit of the ``Three Laws of Robotics'') and for endowing it with a sense of self-awareness and self-assertion that can make it appear more animal-like and autonomous.

Facial expression has been a dominant channel for emotion display, and many social robots leverage facial cues [1-3]. However, clearly legible facial expressions are not universal across species. By contrast, motion-based nonverbal displays span multiple levels of meaning, including animal-like signals and instinctive forms of rejection or aversion. In this work, we design a system that expresses the evolving internal state of dislike toward an approaching other through simple body motions.

Expressing such instinctive internal states through motion can increase perceived animacy [4] and lifelikeness [5], shifting the robot from merely ``friendly'' to a more agentic entity with will and affect. To capture vigilance and dominance/submission dynamics, we vary behavior patterns along the Dominance axis of the PAD model [6].


\section{System Design}
The proposed system generates robot motions for both the endurance phase---i.e., the build-up until an explosion---and the avoidance motion executed at the moment of that explosion. To this end, it computes internal parameters that vary with a counterpart's approach and maps those parameters to expressive, emotion-like motions. In addition, we consider the social relationship with the counterpart and the robot's Dominance tendency (personality) to derive distinct behavior patterns. Accordingly, our internal-parameter design includes: (i) quantifying the state via an accumulation model of dislike; (ii) setting a tolerance threshold that defines the endurance limit; and (iii) a behavior-generation module that expresses different action patterns as a function of Dominance, as detailed below.

Concretely, the design comprises three parts:
\begin{enumerate}
  \item \textbf{Internal state based on accumulation of dislike.} We quantify momentary dislike elicited by approach and accumulate it over time with decay, capturing build-up during endurance.
  \item \textbf{Thresholding for tolerance and limit-exceeding.} A tolerance threshold scales endurance-motion intensity; crossing a limit triggers a one-shot avoidance action whose intensity reflects the excess.
  \item \textbf{Behavior generation modulated by Dominance and relationship.} Motion repertoires and gains depend on Dominance (low/medium/high) and relationship (e.g., stranger, acquaintance, friend, partner).
\end{enumerate}

\subsection{Internal State and Parameters}
\textbf{Momentary dislike:} estimated from approach (e.g., distance and its change). 
\textbf{Accumulation with decay:} the internal state integrates momentary dislike with a forgetting factor, modeling gradual build-up and relaxation.

\subsection{Tolerance and Avoidance Logic}
\textbf{Endurance phase:} while below the tolerance threshold, the robot exhibits endurance motions whose amplitude or tempo scales with the internal state. 
\textbf{Explosion trigger:} when first exceeding the threshold (up to a preset limit), the robot emits a one-shot avoidance action with intensity proportional to the excess.

\subsection{Dominance- and Relationship-Dependent Patterns}
\textbf{Dominance modulation:} 
low--subdued endurance, avoidance as escape; 
medium--noticeable endurance, avoidance as push-away; 
high--restless endurance, avoidance as assertive strike-like motion (within safety constraints).
\textbf{Relationship tuning:} thresholds and motion gains are adjusted to reflect social context.

\begin{figure}[tb]
	\begin{center}
        \includegraphics[width=\hsize]{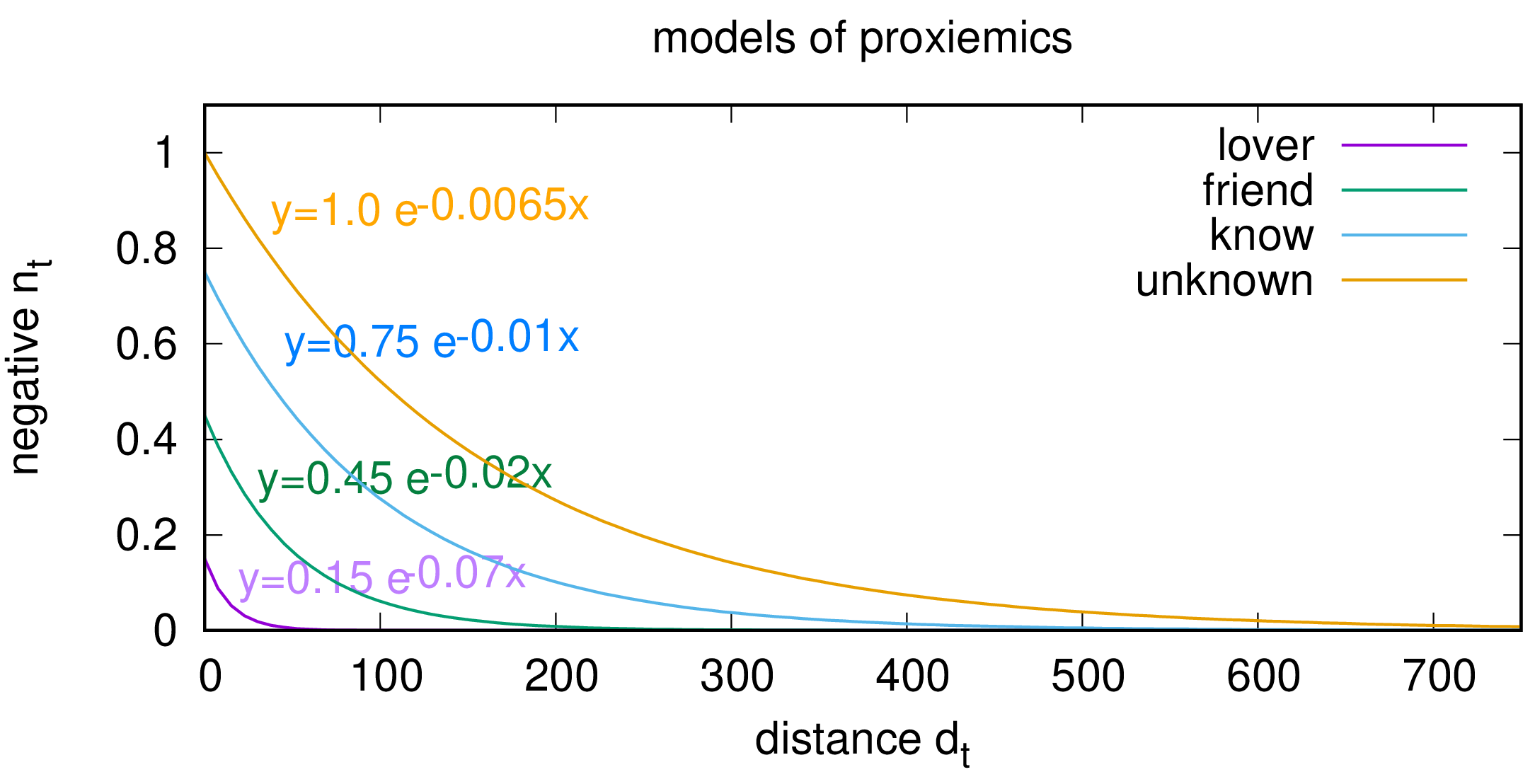}
        \vspace{-3mm}
	\caption{Dislikeness model correspoinding to distance}
	\label{fig:ndistance}
	\end{center}
\end{figure}

\begin{figure*}[tb]
	\begin{center}
    \includegraphics[width=0.8\textwidth]{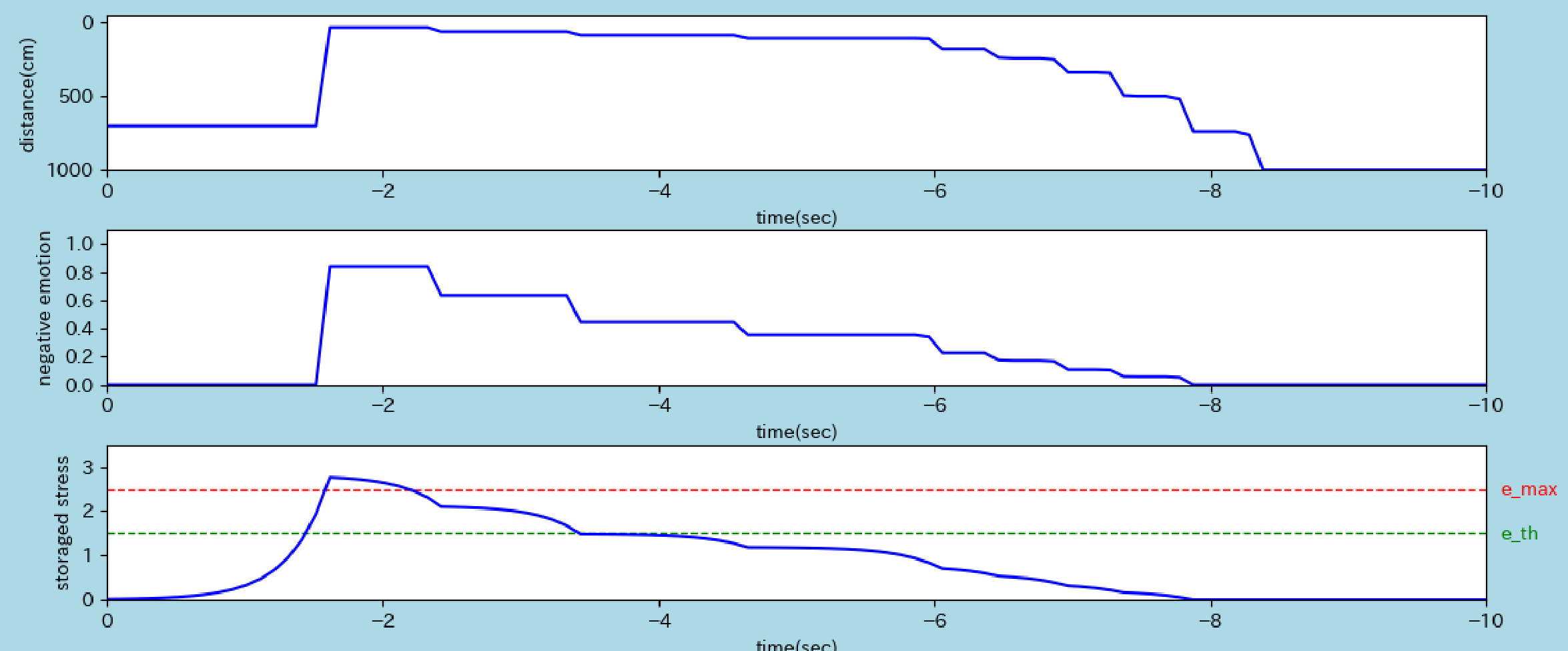}\\
	\caption{Time-series model of dislike accumulation during endurance and the subsequent explosion.
In this graph, the horizontal $t$-axis uses the current time as 0. As time passes, the plotted data flows rightward.	} 
	\label{fig:nmodel}
	\end{center}
\end{figure*}

\begin{table}[tb]
	\begin{center}
	\caption{Endurance motions and avoidance behavior patterns with \textit{Dominance}}
	\label{table:mopattern}
	\begin{tabular}{c c c}
	\hline
	 \textit{\bf Dominance} & \bf Endurance motion & \bf Avoidance motion\\
	\hline
Low    & Slumping                         & Escape  \\
Medium & Deep breathing                   & Push away \\
High   & Jitter (leg-jiggling-like)       & Strike \\	
	\hline
	\end{tabular}
	\end{center}
\end{table}

\begin{figure*}[tb]\vspace{1mm}
	\begin{center}\begin{scriptsize}
\begin{tabular}{cccp{3cm}}
Motion types&Phase 1& Phase 2& \\ 
    \begin{picture}(0,0)(0,0)\put(-5,30){\scriptsize Escape}\end{picture}&
    \includegraphics[width=0.23\textwidth]{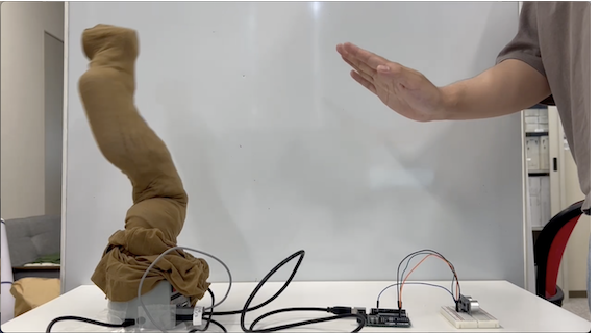}&
    \includegraphics[width=0.23\textwidth]{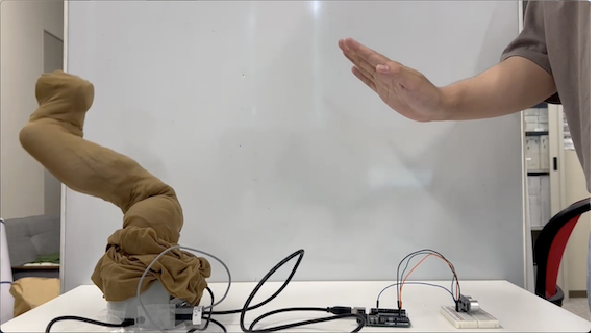}&
 \begin{picture}(0,0)(0,0)\put(0,40){\scriptsize The robot arm reacts to the }\put(0,30){\scriptsize closer distance with the user's}
 \put(0,20){\scriptsize hand than the threshold.}\end{picture}
    \\ 
    \begin{picture}(0,0)(0,0)\put(-5,30){Push away}\end{picture}&
    \includegraphics[width=0.23\textwidth]{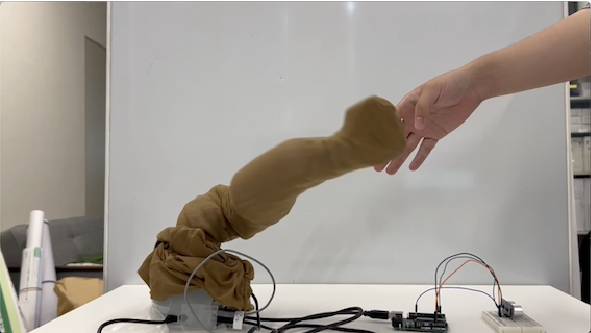}&
    \includegraphics[width=0.23\textwidth]{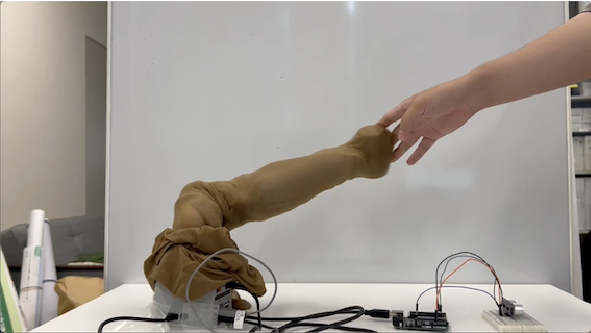}&
 \begin{picture}(0,0)(0,0)\put(0,40){\scriptsize The robot pushes the user's} \put(0,30){\scriptsize hand to the right side, away.}\end{picture}
    \\
    \begin{picture}(0,0)(0,0)\put(-5,30){Strike}\end{picture}&
        \includegraphics[width=0.23\textwidth]{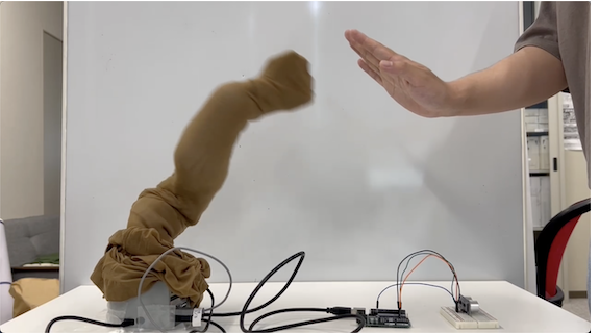}&
    \includegraphics[width=0.23\textwidth]{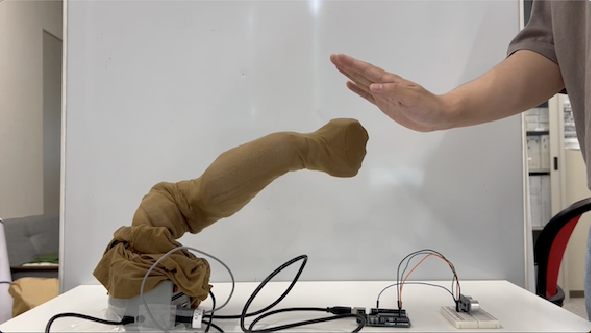}&
 \begin{picture}(0,0)(0,0)\put(0,40){\scriptsize The robot straightly streches} \put(0,30){\scriptsize its arm to the user's hand}\end{picture}
    \\
\end{tabular}    
   \end{scriptsize}

	\caption{Avoidance motion set}
	\label{fig:examotion}
	\end{center}
\end{figure*}
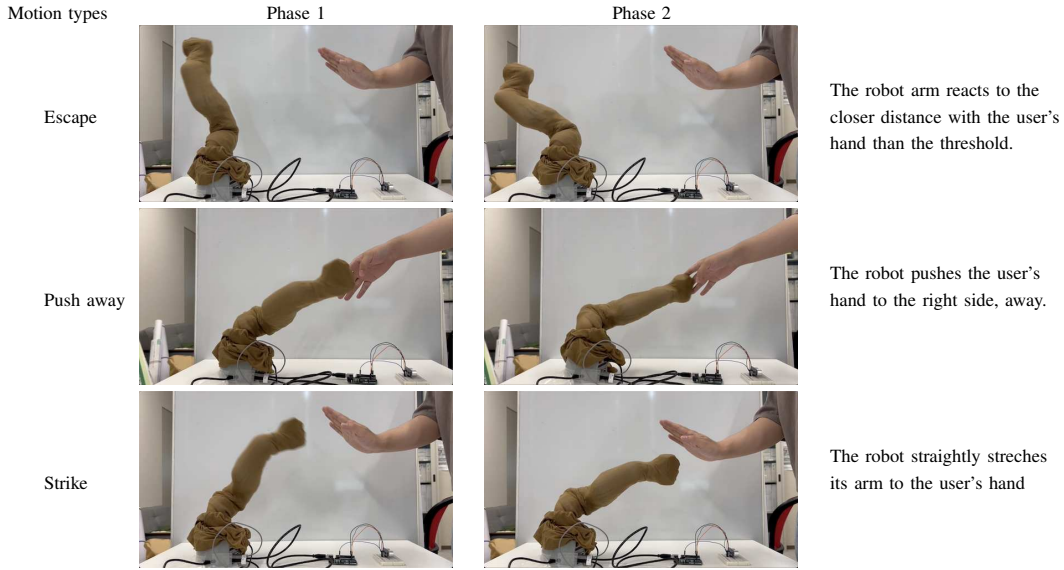

\begin{figure}[htb]
	\begin{center}
    \includegraphics[width=0.25\textwidth]{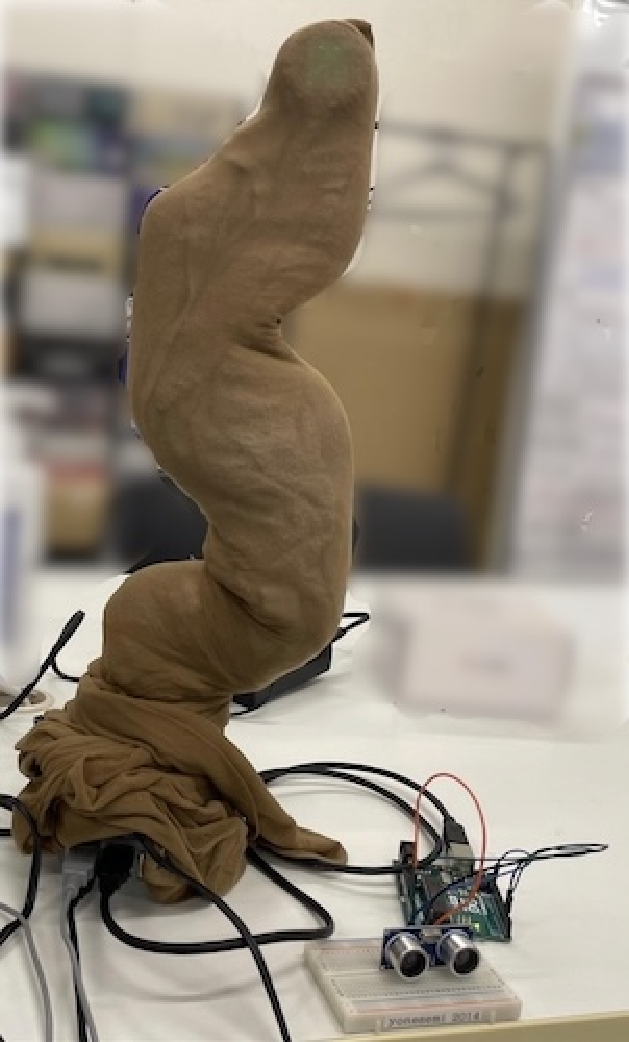}\\
	\caption{Hardware structure.
	Ultrasonic distance sensor (bottom right) connected to the myCobot via an Arduino.
	The myCobot is covered with a stocking/stretch fabric to prevent a robot-like appearance.
	}		
	\label{fig:hardware}
	\end{center}
\end{figure}

\subsection{Internal Parameter Design}

\paragraph{\textbf{Computing dislike at time $t$}}
When a human approaches closer than the robot's preferred personal space, we define the amount of momentary \textit{dislike} at time $t$ (in frames) as a function of the distance $d_t$. Let this be $n_t$, and set
\[
n_t = a \cdot e^{b \cdot d_t},
\]
where $a$ and $b$ are constants. Following Hall's proxemics~\cite{hall}, we provisionally assign numerical ranges based on the near/far bounds of the four zones (intimate, personal, social, public) and fit the above form accordingly. The resulting approximation is shown in Fig.~\ref{fig:ndistance}.

\paragraph{\textbf{Decay and accumulation of dislike up to time $t$}}
Let $s_{t-1}$ denote the accumulated dislike just prior to time $t$. We integrate past values with a decay factor $c \in [0,1]$ over time:
\[
s_t = n_t + s_{t-1} \cdot c
\quad\text{(equivalently } 
s_t=\sum_{i=0}^{t} n_i \cdot c^{\,t-i}\text{).}
\]
Figure~\ref{fig:nmodel} depicts the model of endurance accumulation, the internal state when the tolerance threshold is exceeded, and the ensuing avoidance behavior. In our prototype, distance measurements are sampled at 0.1~s per frame (10~fps); we increment $t$ at each sample and update $s_t$. When the tolerance threshold is crossed, an \textit{avoidance} action is triggered; while the robot is still enduring (below threshold), it continuously displays \textit{endurance} motions with intensity scaled to the current level of dislike.

\paragraph{\textbf{Thresholds for the intensity and timing of endurance/avoidance}}
To determine when behaviors occur and how strong they are, we set two thresholds based on (i) the relationship with the user and (ii) the robot's personality: the \textit{tolerance threshold} $e_{\mathrm{th}}$ and the \textit{maximum admissible dislike} $e_{\max}$. While $s_t \le e_{\mathrm{th}}$, the robot exhibits \textit{endurance} with intensity proportional to $\frac{s_t}{e_{\mathrm{th}}}$. The moment $s_t > e_{\mathrm{th}}$, it emits an \textit{avoidance} action with intensity proportional to $\frac{s_t}{e_{\max}}$.

For example, with a familiar friend, it is easier to display casual rejection while actual tolerance is higher; thus $e_{\mathrm{th}}$ tends to be \textit{lower} and $e_{\max}$ \textit{higher}. In contrast, for an acquaintance or someone with whom it is hard to express true feelings, $e_{\max}$ is \textit{lower} than in the friend case, and the time to reach the limit becomes longer; consequently, the gap between $e_{\mathrm{th}(\mathrm{acq})}$ and $e_{\max(\mathrm{acq})}$ becomes \textit{smaller}.

\subsection{Robot Motion Generation}
In this prototype, we focus on expressing \textit{avoidance} behaviors with an arm robot. Because an arm robot differs from a humanoid in degrees of freedom, the current system uses design-preset motion patterns prepared in advance per \textit{Dominance} level. As shown in Table~\ref{table:mopattern}, we prepared three \textit{endurance} motions and three \textit{avoidance} actions according to Dominance. Examples of the avoidance motions are illustrated in Fig.~\ref{fig:examotion}.

The hardware consists of an HC-SR04 ultrasonic distance sensor, an Elephant Robotics myCobot 280 Pi (6-DOF) collaborative arm covered with stretch fabric to mask a robot-like appearance, and an Arduino~UNO. Sensor readings are sent via serial communication to the myCobot, where the internal state is computed. The HC-SR04 is provisionally mounted facing the user's approach direction. Due to its specifications, distances are only estimated between 2~cm and 450~cm; hence we clamp values below 2~cm to 0~cm and above 450~cm to 700~cm.

When $n_t>0$, the system enters the \textit{endurance} phase: using motion interpolation, it continuously expresses an endurance motion from a neutral waiting pose. The intensity is scaled by $\frac{s_t}{e_{\mathrm{th}}}$, which modulates the motion amplitude (e.g., angle change). When $s_t>e_{\mathrm{th}}$, the robot performs an \textit{avoidance} action (escape, push away, or strike-like motion). The action speed is scaled by $\frac{s_t}{e_{\max}}$. This realizes continuous-intensity \textit{endurance} below the limit and graded-intensity \textit{avoidance} once the limit is exceeded.

\subsection{Example Performance Flow}
In the current implementation we do not estimate the person with a camera; instead, prior to interaction we select one of \{stranger, acquaintance, friend, partner\} for the user the robot will face. Based on this label and the current distance $d_t$, we compute $n_t$.

\textbf{Friend, $d_t=30$~cm.} From the ``friend'' approximation in Fig.~\ref{fig:ndistance}, $n_t \approx 0.25$. If the hand remains at that distance, the robot keeps integrating with decay to obtain $s_t$, and displays endurance with increasing intensity. With \textit{high Dominance} toward a friend, the endurance motion is \textit{jitter}; with \textit{medium Dominance}, \textit{deep breathing}; both scaled by $\frac{s_t}{e_{\mathrm{th}(\mathrm{friend})}}$. With $c=0.7$ and $e_{\mathrm{th}(\mathrm{friend})}=0.75$, the threshold is crossed at $t=7$ frames, triggering an avoidance action with intensity $\frac{s_t}{e_{\max(\mathrm{friend})}}$; at 10~fps this corresponds to 0.7~s. If $d_t=10$~cm, the crossing occurs at $t=3$ frames (0.3~s). Thus, with a friend the robot shows \textit{dislike} relatively quickly. In this case, \textit{high Dominance} yields a \textit{strike-like} avoidance; \textit{medium Dominance} yields \textit{push away}. Prolonging the unpleasant state increases $s_t$ and strengthens avoidance intensity.

\textbf{Acquaintance.} At $d_t=30$~cm, $n_t \approx 0.56$. If distance is maintained and $e_{\mathrm{th}(\mathrm{acq})}=2.0$, $s_t$ saturates at 1.85, so the robot continues \textit{endurance} and does not show avoidance. At $d_t=20$~cm, avoidance occurs at $t=11$ frames; at $d_t=10$~cm, at $t=7$ frames. With \textit{medium Dominance} the avoidance is \textit{push away}; with \textit{low Dominance} it is \textit{escape}.

These differences---how \textit{endurance} accumulates, where the limit lies, and which \textit{avoidance} pattern is selected---allow the robot to communicate its history of enduring. We therefore set $e_{\max(\mathrm{friend})}$ \textit{higher} and $e_{\mathrm{th}(\mathrm{friend})}$ \textit{lower}. For acquaintances, $e_{\max(\mathrm{acq})}$ is \textit{lower} and $e_{\mathrm{th}(\mathrm{acq})}$ \textit{higher}, reflecting how easy it is to express \textit{dislike} in that relationship.

\section{Future Work}
This prototype uses a single ultrasonic sensor and assumes a fixed hand position relative to the torso, without fine-grained position estimation for aggressive avoidance. It is therefore not robust to diverse approach trajectories. To increase range and accuracy, we will consider replacing or combining with IR or other sensors, or introducing camera-based ranging. A camera would also enable user identification and hand localization, allowing avoidance that does not depend on the torso's position.

As a time-series process, the current implementation yields relatively quick avoidance responses. We plan to examine long-term endurance parameters for cases where near-threshold dislike is repeatedly induced without crossing the limit.

The accumulation model requires validation to determine appropriate constants, and personal adaptation to handle variability. Since $c$, $a$, and $b$ may vary with the robot's personality, we ultimately need mappings from relationship to the parameters of $n_t$ ($a,b$), the thresholds $e_{\mathrm{th}}, e_{\max}$, and the Dominance setting, so that relationships can be handled on a continuous scale.


Here, as discussed in prior work about aggressive and joking attitudes~\cite{maehama21}, it can be difficult to distinguish the robot's intentional from unintentional expressions. Beyond physical or psychological self-defense~\cite{duarte22}, people sometimes deploy \textit{tatemae} (a socially appropriate public stance) as a form of social defense---effectively an avoidance of another's presence at the level of social appearance rather than essence. In contrast, \textit{honne} denotes private, genuine feelings. We argue that avoidance behaviors should be designed with \textit{honne}-\textit{tatemae} dynamics in mind---i.e., the robot's public face versus its private state---grounded in relationship and social context~\cite{yoshida24}. Otherwise, revealing a robot's ``real feelings'' too bluntly can distress users and reduce their willingness to live with a robot perceived as aggressive \emph{when the relationship does not yet feel close to the user or when others are present and watching}.

\section{Ethics and Regulation}
We proposed an avoidance-behavior framework for robots' expression of refusal and self-protection. However, harmful striking would violate the spirit of the Three Laws. Implementations must ensure non-painful contact for avoidance actions (push away / strike-like), with safety constraints tuned to mechanical structure and stiffness.

Avoidance may also cause negative affect, anxiety, or pain for users. Ethically, designers must consider potential psychological harm from being rejected---even by a robot. Rejection from a familiar partner robot can be particularly shocking. Users should be informed about the robot's personality and behaviors beforehand, and settings should adapt to the user's mental state. Clear rules are needed to balance the robot's right to avoid with the user's physical and psychological safety.

\section{Conclusion}
We presented models for the accumulation and decay of \textit{dislike} in response to human physical approach, together with an \textit{avoidance} model. By setting relationship-dependent dislike parameters, endurance thresholds, and \textit{Dominance}, the robot can display different avoidance behaviors with appropriate timing and intensity. Future work will evaluate the models to determine constants and examine how avoidance expression contributes to a robot's perceived mind and lifelikeness.

\section*{Acknowledgments}
This work was partially supported by JSPS KAKENHI 23K11278, 23K11202, JST CREST JPMJCR18A1, and JST Moonshot R\&D JPMJMS2215 (items related to safety and ethical aspects).


\end{document}